# Automated Low-cost Terrestrial Laser Scanner for Measuring Diameters at Breast Height and Heights of Forest Trees

**Pei Wang** [1,*], **Guochao Bu** [1], **Ronghao Li** [1], **Rui Zhao** [1]

[1] Institute of Science, Beijing Forestry University, No.35 Qinghua East Road, Haidian District, Beijing, China; wangpei@bjfu.edu.cn

* Correspondence: wangpei@bjuf.edu.cn; Tel.: +86-10-6233-8136

**Abstract:** A terrestrial laser scanner is a kind of fast, high-precision data acquisition device, which had been applied more and more to the research area of forest inventory. In this study, a type of automated low-cost terrestrial laser scanner was designed and implemented based on a SICK LMS-511 two-dimensional laser radar sensor and a stepper motor. The new scanner was named BEE, which can scan the forest trees in three dimension. The BEE scanner and its supporting software are specifically designed for forest inventory. The specific software was developed to smoothly control the BEE scanner and to acquire the data, including the angular data, range data, and intensity data, and the data acquired by the BEE scanner could be processed into point cloud data, a range map, and an intensity map. Based on the point cloud data, the trees were detected by a single slice of the single scan in a plot, and the local ground plane was fitted for each detected tree. Then the diameters at breast height (DBH), tree height, and tree position could be estimated automatically by using the specific software. The experiments have been performed by using the BEE scanner in an artificial ginkgo forest which was located in the Haidian District of Beijing. Four $10\,m \times 10\,m$ square plots were selected to do the experiments. The BEE scanner scanned in the four plots and acquired the single-scan data, respectively. The DBH, tree height, and tree position of trees in the four plots were estimated and analyzed. For comparison, manually-measured data was also collected in the four plots. The trunk detection rate for all four plots was 92.75%; the root mean square error of the DBH estimation was 1.27 cm; the root mean square error of the tree height estimation was 0.24 m; and the tree position estimation was in line with the actual position. Experimental results show that the BEE scanner can efficiently estimate the structure parameters of forest trees and has good potential in practical applications of forest inventory.

**Keywords:** laser scanning; terrestrial laser scanning; low-cost; diameter at breast; tree height; tree position

**PACS:** J0101

## 1. Introduction

Laser scanning is a surveying method which can measure the distance between the laser scanner and the point on the object illuminated by a laser. In the past decades, laser scanning technology has been widely used in various fields, such as mapping [1,2], photography [3], surveying [4], and so on. Airborne and spaceborne laser scanning systems have been also used to acquire the point cloud data of large areas of forest and create a forest inventory. The airborne and spaceborne point cloud data, alone or integrated with other kinds of remote sensing data, like optical satellite image and SAR (synthetic aperture radar) images, are often used to characterize forest canopies [5], estimate the crown base height [6], estimate forest biomass [7], and capture tree crown formation [8].

Terrestrial laser scanning (TLS) is different from airborne and spaceborne laser scanning. TLS is often used to acquire the detailed data of objects in a local area. TLS data can describe the forest in detail with very high accuracy even to the millimeter level. Without doubt, TLS data has the potential to perform the plot-level or tree-level forest inventory at very high precision. Thus, TLS was used to collect the spatial information of the forest and measure the forest structure [9,10], and more and more experiments were carried out to estimate the attributes of the forest trees, such as the diameter at breast height (DBH) [10,11], tree height [12], tree crown [13], leaf area index [14], and biomass [15], etc..

Among these studies, one focus was the accurate estimation of the structure parameters, such as the DBH and tree height. The estimation of DBH in point cloud data is mostly based on circle fitting. The trunk is treated as a cylinder, then transects of the point cloud data at breast height were used to determine DBH. Simonse [16] and Linderg [17] used a 2D Hough transform to detect trees and estimate the DBH. Other circle fitting methods were used in the estimation of the DBH. Generally, there are two kinds of circle fitting methods. One kind is the geometrical method, such as the Gauss-Newton method and the Levenberg-Marquardt method. The other is the algebraic method, such as the Pratt method [18] and Taubin method [19]. In practice, the algebraic methods are non-iterative and faster than the geometric methods. For general data, these circle fitting methods are not very different in terms of accuracy for DBH estimation. Pueschel studied the circle fitting methods and thought that removal of outliers in data is very important in circle fitting [11].

Since the DBH is defined as the girth at 1.3 m above ground level, the other factor affecting the accuracy of DBH estimation is the ascertainment of the ground level. Watt derived the ground level of individual tree by manually rotating the point cloud data and observing from several view angles [9]. Bienert [20] and Mass [12] determined the ground level by using a digital terrain model (DTM) which was created by using the density allocation along the z-axis. Olofsson used TLS data to construct a digital elevation model (DEM) to define the breast height [21]. Yang also derived a fine-scale DEM with a $0.5 m \times 0.5 m$ grid by using the random sample consensus (RANSAC) algorithm and an interpolating algorithm [22]. If the ground is flat with less understory, the determination of the ground level is relatively easy. Otherwise, if the ground is uneven or with dense understory, the ground level is harder to determine. Furthermore, we do think that creating the DTM or DEM of the sample plot is more time consuming than finding the local ground level for individual trees directly.

When the local ground level is determined, the uppermost point of the tree is the key for total height estimation. Some papers used a cylinder space which surrounds each tree. In the cylinder, the highest point was defined as the uppermost point and the tree height can be calculated [12,20]. Some papers used the tree crown outlines [21] and canopy height model [22] to find the uppermost point and, thus, calculate the tree height. If the trees are leaf-on and the leaves are dense, the laser light will be mostly blocked and the crown shape would be the better method to determine the uppermost point. If the trees are leaf-off and with many branches and twigs, the outliers would affect the selection of the uppermost point significantly and the removal of outliers would be necessary.

By the above description, obviously TLS is a very useful technology for forest inventory. TLS could acquire massive amounts of data of the forest with very high speed and high accuracy. Manually processing the data is very time consuming and inaccurate, so automated processing becomes a trend. Many methods have been proposed to automatically estimate those structure parameters [12,23–26].

However, there are still problems regarding the forest applications of the TLS data. One problem is that the forest inventory experiments are mostly carried out by using expensive commercial 3D terrestrial laser scanners of companies like Leica Geosystems (Heerbrugg, Switzerland), RIEGL (Horn, Austria), FARO (Lake Mary, Florida, USA), etc. These commercial terrestrial laser scanners have higher accuracy, longer range, and more powerful data acquisition capabilities, but the high price increases the cost of widely using TLS data in forest applications. The other problem is that those expensive scanners are common equipment, which are not limited to forestry. These commercial scanners and their software are mostly of a common type, which can be used for mapping, architecture, forestry, archeology, and other fields. They are not designed for forestry applications, and they lack the specialized processing software necessary for forest inventory. However, when the user requires a stable and robust software and hardware platform, more accurate data, and are not sensitive to price, these commercial 3D terrestrial laser scanners are the better choices.

Therefore, automated low-cost terrestrial laser scanners designed for forest inventory are needed. A low-cost scanner is conducive to more researchers entering this area of study and allow more research on forest applications. A variety of types of scanners can provide more options and stimulate more innovative ideas on forestry applications.

Some related work has been done in this area. Jaakkola designed a low-cost multi-sensor mobile mapping system by using a mini-UAV as the platform [23]. In that system, two laser scanners were used to obtain the point cloud data. One scanner was an Ibeo Lux and the other was a SICK LMS151, and many sensors were integrated into the system, such as a GPS/IMU positioning system, a CCD camera, a spectrometer, and a thermal camera. The system was used to measure tree height and derive the biomass change of a coniferous tree. Kelbe designed a low-cost terrestrial laser scanner with a SICK LMS-151 system [24]. The system was tested in a $20\,m \times 20\,m$ plot, and the point cloud data of a single scan was used to reconstruct the 3D tree stem models and estimate the DBH and tree position. Liang designed a personal laser scanning (PLS) system with a FARO Focus3D 120 for forest mapping and ecosystem services [25]. They proposed a multipass-corridor-mapping method based on the system and did experiments in a 2000 $m^2$ forest plot. The system could map the test plot in two minutes and estimate the DBH accurately. Kong proposed a new hybrid algorithm for estimating the DBH by using a SICK LMS-511 and a thermal camera [20]. In that work, two-dimensional scanning points collected by multiple scans were used to estimate the DBH. Culvenor designed an automated in situ laser scanner for monitoring the forest leaf area index [27]. Olofsson developed the software tools for automatic measurement and analysis based on point cloud data in C, Python, and the R programming languages [21].

This paper describes an automated low-cost 3D terrestrial laser scanner for measuring the DBH, tree height and tree position of the local forest. The scanner described in the paper is composed of a SICK LMS-511 and a stepper motor. The scanner was tested in four plots in the experiments. The acquired point cloud data of trees were processed by a series of processes, such as the trunk detection, ground fitting, DBH estimation, tree height estimation, etc., and the estimated results of the plots were analyzed and discussed to demonstrate the feasibility and the potential of the automated low-cost terrestrial laser scanner.

## 2. Materials and Methods

### 2.1. Instrument Description

The goal of this study was to develop an automated low-cost terrestrial laser scanning system for measuring DBH, tree height, and tree position. The developed scanner is named BEE, which stands for the developers that are working or studying at the **B**eijing Forestry University, and majoring in **E**lectronic **E**ngineering. The BEE scanner is composed of two main parts, hardware and software. By selecting and developing the key components of the hardware, such as the laser scanning sensor and the platform, the cost of the system was below $10,000. The laser scanning sensor implements the emission and the reception of laser light. In this study, a two-dimensional laser scanning sensor was used. The platform implements the rotation of the laser scanning sensor and the communication between components. A stepper motor was used to rotate the laser scanning sensor to fulfill the three-dimensional scanning.

By considering the performance and price of the laser scanning sensor, we chose the SICK LMS-511 laser scanning sensor, which is manufactured by the SICK company (Waldkirch, Germany). Table 1 lists the technical specifications of the SICK LMS-511 laser scanner. The laser scanning sensor employs a class 1 laser that operates in the near infrared part of the spectrum at 905 nm. The angular step-width could be preset at several values, such as 0.1667°, 0.25°, 0.333°, 0.5°,

0.667°, and 1°. The scan speed could also be preset at several values, such as 25 Hz, 35 Hz, 50 Hz, 75 Hz, and 100 Hz. The angular coverage of the laser scanning sensor is up to 190°, from –5° to 185°. The laser scanning sensor's power requirements are supplied by a 24 V DC supply. The laser scanning sensor can measure echoes from the object's points. The measured range and intensity data can be recorded remotely via an Ethernet interface or a RS232 interface to the external computer.

**Table 1.** Specifications of the SICK LMS-511.

| Performances | Specifications |
|---|---|
| Measurement Distance(m) | 0.7–80 |
| Supply Voltage(V) | 24 V DC ± 20% |
| Accuracy(mm) | ±12 |
| Laser Wavelength(nm) | 905 |
| Scan Speed(Hz) | 25, 35, 50, 75, 100 |
| Angular Step width(°) | 0.167, 0.25, 0.333, 0.5, 0.667, 1 |
| External Dimensions(mm) | 160 × 155 × 185 |

The laser scanning sensor was mounted on a rotating platform which is set horizontally to the ground, and the scanning laser sensor was perpendicular to the ground so that the angular coverage is 190° in the vertical direction. The platform implements 360° scanning in the horizontal direction by using a motor which is also supplied by 24 V DC. Therefore, the field of view of the BEE scanner is 360° × 190°.

The specific software, named BEEScan, was developed for controlling the BEE scanner, receiving the data from the scanner, estimating the structural parameters of trees automatically, and displaying the results. BEEScan was developed by using C, C++, and Python programming languages. BEEScan contains three modules: a ScanController module, a Display module, and TreeStructureExtract module, as shown in Figure 1. The ScanController module controls the components of the BEE scanner to work together harmoniously, receiving and recording the measured data on a laptop computer. The Display module demonstrates the 3D effect of the measured data, and the TreeStructureExtract module automatically detects the trees in the scene and measures the DBH, height, and position of the trees.

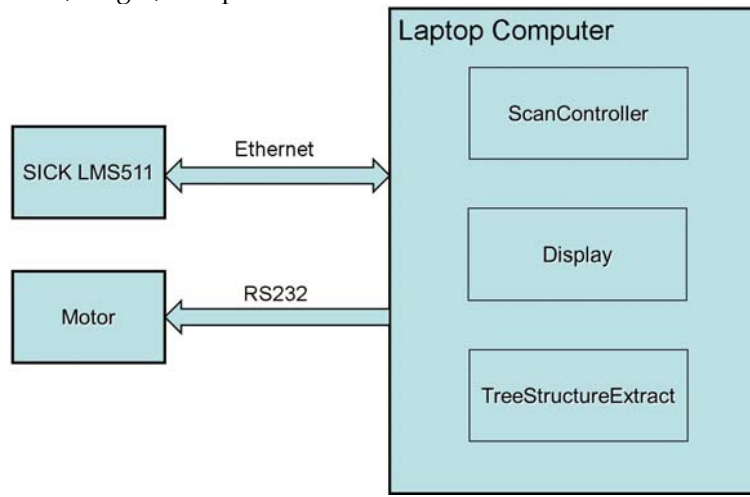

**Figure 1.** The scheme map of the BEE scanner.

The BEE scanner was mounted on a tripod in the field measurement, and connected to a laptop computer, as shown in Figure 2. The BEE scanner was equipped with a 24 V DC battery pack to supply the laser scanning sensor and the stepper motor and a solar battery also was provided as a standby power source.

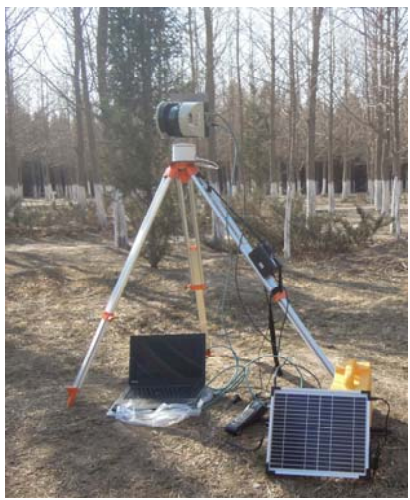

**Figure 2.** Picture of the BEE scanner at the field site.

*2.2. Study Area*

The study area is located in a man-made mixed plantation forest in the Haidian District of Beijing (37.7° S, 144.9° E), Figure 3a. The forest trees are mainly ginkgo trees and a small number of pine trees, Figure 3b. The planting space between trees is about 1.6 m in the north-south direction, and 2.8 m in the west-east direction. Some trees died and were removed, and their locations were vacant. The stem density of the forest is about 0.2 trees/m². The mean and standard deviation are 12.11 cm and 2.37 cm for DBH and 8.43 m and 0.78 m for tree height. The experiment was carried out in March 2015 when the ginkgo trees were in the leaf-off period.

In the forest, four $10\,m \times 10\,m$ square plots were selected for the scanning experiments. Plots of $10\,m \times 10\,m$ were studied based on the following considerations: Firstly, in the experiments, DBH and tree height were estimated at the same time based on the acquired point cloud data. The BEE scanner has the angular resolution of 0.167 degrees in the vertical direction. The interval between points in the vertical direction is about 30 cm at a distance of 10 meters away from the scanner. If the size of the plot becomes larger, the interval will become larger, which is not favorable to estimate the tree height. Secondly, when the size of the plot becomes larger, the trees far away from the scanner are scanned less. Thus, the point cloud data of these trees are mostly sparse and incomplete, which is also not favorable to estimate the DBH and tree height.

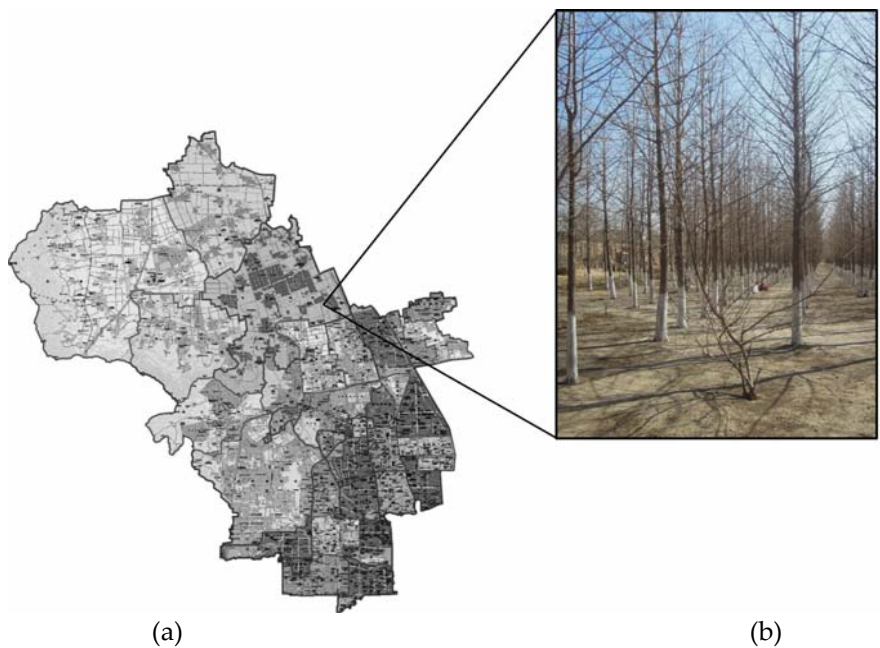

(a)          (b)

**Figure 3.** (**a**) The study area located in Haidian district of Beijing (37.7° S, 144.9° E); and (**b**) a picture of the field site.

*2.3. Data Acquisition*

Four $10\,m \times 10\,m$ square plots were selected to be scanned in the experiments. The scan position was selected near the center of each plot considering a good field of view, which could reduce the shadowing effects. The BEE scanner was mounted on a tripod, which was placed at the selected positions. The rotating speed was set one degree per second so that the BEE scanner could scan the 360° scene in six minutes. When the BEE scanner started working, the dataset was transferred to the laptop computer and recorded. A single scan was performed in each plot.

For comparison, the DBH of each tree in the four plots was tape-measured and recorded. Tree height could be deduced according to the distances and angles to the top and bottom of the tree, which were acquired by using a Bushnell hand-held rangefinder. Tree heights were also calculated and recorded, and each tree in the plots was numbered and its relative position in the plot was drawn. The above manual measured data would be used for the consecutive data processing and comparison with the estimated values.

## 3. Methods

When the BEE scanner was settled down in the plot, the scan was started by using the designed software. The system would run step by step as shown in Figure 4. A large amount of data, including the horizontal angle, the vertical angle, the intensity data, and the range data of each scanning point was obtained by the computer. The calculated point cloud data was used to detect the trees in the plot and fit to the local ground plane of each tree. Then the DBH, tree height, and tree position were estimated consecutively.

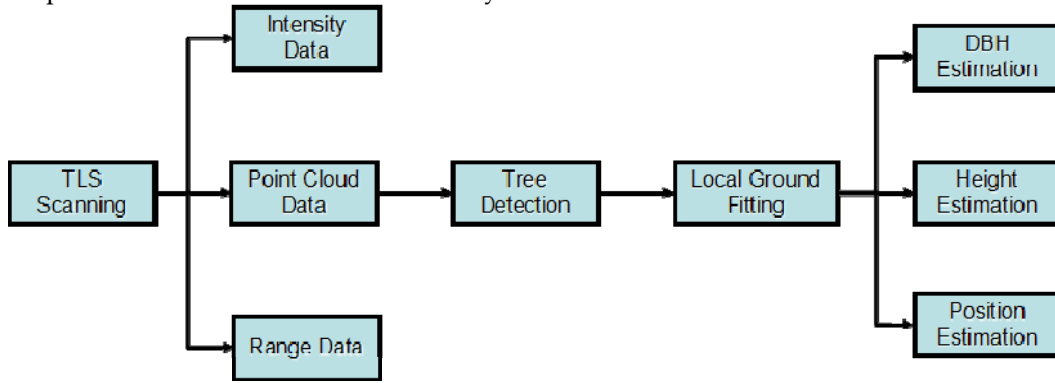

**Figure 4.** The flowchart of the data acquisition and processing of the BEE scanner.

*3.1. Acquiring the Scanning Data*

The SICK LMS-511 is a type of two-dimensional laser scanning sensor which can scan the contour of measured objects by using the time of flight of the laser pulse. The slant range $R$ and the intensity value $I$ of each measured point were provided to the user. Meanwhile, the user could calculate the vertical angle $\theta$ of each point according to the preset vertical angular step width which is set to 0.1667° in the experiments. When the SICK LMS-511 was placed vertically and rotated by the stepper motor, the horizontal angle $\varphi$ of each point can be acquired by counting the steps of the stepper motor at a certain speed of rotation. Thus, there is a set of information $(R, I, \theta, \varphi)$ for each measured point.

Take the scanning of trunks as an example, which is shown in Figure 5. The trunk was plotted as a cylinder. $S$ denotes the BEE scanner. The scanner measures the slant range $R$ of each sampled point on the trunk, and records the vertical angle $\theta$ and the horizontal angle $\varphi$. Point $A$ and point $C$ are adjacent in the vertical direction, and the vertical angular step width is $\Delta\theta$.

Point $A$ and point $B$ are adjacent in the horizontal direction, and the horizontal angular step width is $\Delta\varphi$.

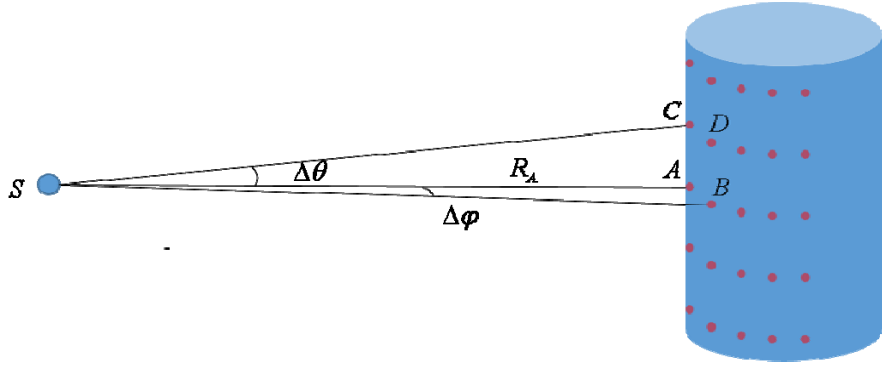

**Figure 5.** The geometrical working diagram of the BEE scanner.

Assuming that the position of the SICK sensor is in the position with the coordinates $(0,0,0)$ in three-dimensional space, the coordinates $(x, y, z)$ of each point could be calculated by using a set of information $(R, \theta, \varphi)$ according to the simple spatial geometrical relationships. For example, when the measured point is in the first quadrant, the coordinates were calculated by:

$$\begin{aligned} x &= R \cdot \cos(\theta) \cdot \cos(\varphi) \\ y &= R \cdot \cos(\theta) \cdot \sin(\varphi), \\ z &= R \cdot \sin(\theta) \end{aligned} \quad (1)$$

In this way, the point cloud data of the whole scene could be constructed in three-dimensional space. These measured points could also be projected onto a cylinder with their range values or intensity values. Then the range map and the intensity map could be provided to the user as the auxiliary data for processing and analysis.

*3.2. Tree Detection*

Since the plots are only $10\,m \times 10\,m$ in size and the ground is relatively flat, the single slice method was used to detect the trees in the point cloud data. We obtained the point cloud transect at a height of 1.3 m by slicing the point cloud data with $\pm 5$ cm height. Then the point cloud transect was projected onto a plane and clustered by using the horizontal and vertical angles and the range [28]. For example, the distances between points adjacent in the horizontal and vertical direction in Figure 5 could be calculated by:

$$\begin{aligned} \text{distance}(A, B) &= d_{AB} = \sqrt{R_A^2 + R_B^2 - 2R_A R_B \cos(\Delta\varphi)} \\ \text{distance}(A, C) &= d_{AC} = \sqrt{R_A^2 + R_C^2 - 2R_A R_C \cos(\Delta\theta)} \end{aligned} \quad (2)$$

That is to say that two points may belong to a same trunk if their distance matched one of the above formulas. However, in actual practice, some factors, like the noise, the irregular shape of the trunk, and errors, must be considered. In the experiments, two points were classified into a cluster when their distance is less than $d_{threshold}$, which is calculated by the following formula, and the value of k was set 1.5 after several attempts:

$$d_{threshold} = k \cdot \max\{d_{AB}, d_{AC}\} \quad (3)$$

If a cluster of points are consistent with the distribution of the circumference in the plane, it means a tree may be detected [11]. Each cluster of points would be used to fit a circle. The center point of the circle seemed to be a tree position when the fitted diameter was in a reasonable range.

BEEScan can provide a variety of circle fitting methods, such as the Gauss-Newton method, Pratt method [18], and Taubin method [19]. The Gauss-Newton method is a geometric fitting method and the other two methods are algebraic fitting methods. The Gauss-Newton method is iterative, therefore the latter two methods are much faster than the Gauss-Newton method. With respect to accuracy, the Gauss-Newton method is only slightly better than the latter two methods, and the difference is only in millimeter. Furthermore, in our practical tests the Pratt method was slightly better than the Taubin method on accuracy and speed. Thus, we used the Pratt method in the practical processing in order to ensure the efficiency while not seriously losing accuracy.

In the Pratt circle fitting method, a circle is described by an algebraic equation and a constraint [29]:

$$A(x^2+y^2)+Bx+Cy+D=0$$
$$B^2+C^2-4AD=1 \qquad (4)$$

The Pratt method aims to find a circle which can minimize the following function and is subject to the constraint:

$$F(A,B,C,D)=\sum_{i=1}^{n}(Az_i+Bx_i+Cy_i+D)^2$$
$$B^2+C^2-4AD=1 \qquad (5)$$

The Taubin method aims to find a circle which can minimize the following function and is subject to the constraint:

$$F(A,B,C,D)=\sum_{i=1}^{n}(Az_i+Bx_i+Cy_i+D)^2$$
$$4A^2\bar{z}+4AB\bar{x}+4AC\bar{y}+B^2+C^2=1 \qquad (6)$$

The Taubin method is similar to the Pratt method, and their performances are similar, as well, which is verified by the practical processing.

The coordinate values of the projected points were used to compute the parameters $(A,B,C,D)$, and then the diameter of the fitted circle could be calculated.

In practice, when we measured the four plots, the thickest stem and the thinnest stem in each plot could be roughly estimated visually, and the rough values could be used as the upper limit and the lower limit, respectively. Thus, a tree position was defined only if the fitted diameter is between the upper limit and the lower limit. If the fitted diameter is out of the range, the circle fitting was thought to use the outliers, and the trunk detection failed.

*3.3. Ground Fitting*

In a large area, the ground may be uneven or fluctuating, and this will affect the judgement of the breast height. Thus, the local ground plane should be fitted first before estimating the DBH. The ground level of the four test plots is relatively flat with less understory. According to the acquired tree position mentioned above, the point cloud data in a cylindrical space is extracted. The cylindrical space is centered on the tree position. The diameter of the cylindrical space is defined according to the tree positions. In the cylindrical space, a small part of the point cloud data was extracted with the lowest height of about 10 cm thickness. Usually the extracted point includes the ground points and the points at the bottom of the trunk. In a local area, the ground is relatively flat. Thus, the local ground plane was fitted by using the RANSAC algorithm [30] in the experiment. If the plane tilt is more serious, the breast height will be corrected. Otherwise, we project the center point of the fitted circle into the fitted ground plane, and the height of the projected point would be the local zero point for the estimation of the tree.

*3.4. DBH, Height, and Position Estimation*

The DBH was measured at breast height of the tree. However, different heights were used in different countries. In China, DBH was measured at a height of 1.3 m. Based on the results of trunk detection and local ground plane fitting, the breast height of 1.3 m could be confirmed. According to the confirmed breast height, the point cloud transect was re-sliced and used to estimate the DBH by using the above mentioned Pratt circle fitting method. Thus, the new center point of the circle was obtained, and the point's projected position on the local ground plane was defined as the local zero-height point $p_{zero}$ and the position of the tree.

Since the trees in the plots are growing straight, and the trees are in the leaf-off period, the total height of a tree would be the vertical distance between the point $p_{zero}$ and the point $p_{top}$ which is the uppermost point of the tree. At the top of the tree, some tree branches are scanned and some are blocked. Small twigs may be scanned in part, or be missed, and there are also some outliers at the top of the tree. We have taken the following steps to find the uppermost point $p_{top}$ and calculate the total height of a tree.

(a) Use $V$ to refer to the cylindrical space of a tree, and $z_{max}$ to denote the z coordinate of the highest point in $V$.

(b) Extract the point cloud data set $V_{top}$. A point $p \in V$ with z coordinate $z_p$ is in $V_{top}$ which must satisfy that:

$$\left| z_{max} - z_p \right| < 0.5m \tag{7}$$

(c) Given $M$ points belong to $V_{top}$, search ten points $\{p_{i1}, p_{i2}, \cdots, p_{i10}\} \in V_{top}$ for each point $p_i \in V_{top}$ by using the k-nearest neighbors (KNN) algorithm [35,36]. The points $\{p_{i1}, p_{i2}, \cdots, p_{i10}\}$ are the ten nearest neighbors of point $p_i$.

(d) Calculate the distances from $p_i$ to $\{p_{i1}, p_{i2}, \cdots, p_{i10}\}$ and use $\{d_{i1}, d_{i2}, \cdots, d_{i10}\}$ to refer to the distances.

$$d_{ij} = \left\| p_i - p_{ij} \right\|, \quad i \in [1, M], \, j = 1, 2, \cdots, 10 \tag{8}$$

(e) Calculate the mean value $md_i$ of the distances $\{d_{i1}, d_{i2}, \cdots, d_{i10}\}$ of each point $p_i \in V_{top}$.

$$md_i = \frac{1}{10} \sum_{j=1}^{10} d_{ij}, \quad i \in [1, M] \tag{9}$$

(f) Calculate the mean value $md$ of the $\{md_1, md_2, \cdots, md_M\}$:

$$md = \frac{1}{M} \sum_{i=1}^{M} md_i \tag{10}$$

(g) Use $md$ to cluster the points in $V_{top}$. If $md_i$ is less than, or equal to, $md$, the point $p_i$ is considered as a tree point. Otherwise, if $md_i$ is greater than $md$, the point $p_i$ is considered as an outlier or a point with low confidence.

(h) Select the point with highest height in the tree point set, and use $p_{top}$ to refer to the point.

(i) Calculate the tree height, which is the vertical distance from the point $p_{top}$ to $p_{zero}$.

## 4. Results

*4.1. Scanning Data*

Since the measuring range of the laser scanning sensor can be up to 80 meters, the point cloud data of a large area will be obtained in a single scan. A field measurement of a single scan is shown in Figure 6, which is colored by using the intensity data. Obviously, the trees are planted almost in lines from the north to the south and from the east to the west.

As shown in Figure 6, the point cloud data around the scanner shows higher density than the point cloud data far away from the scanner. The density of the point cloud data varied because of the diffuseness, and the trees far away from the scanner were partially scanned or not scanned because of the shadowing effects of the laser light. As can be seen from the figure, a single scan is not suitable to estimate the DBH and tree height at the same time in a large area.

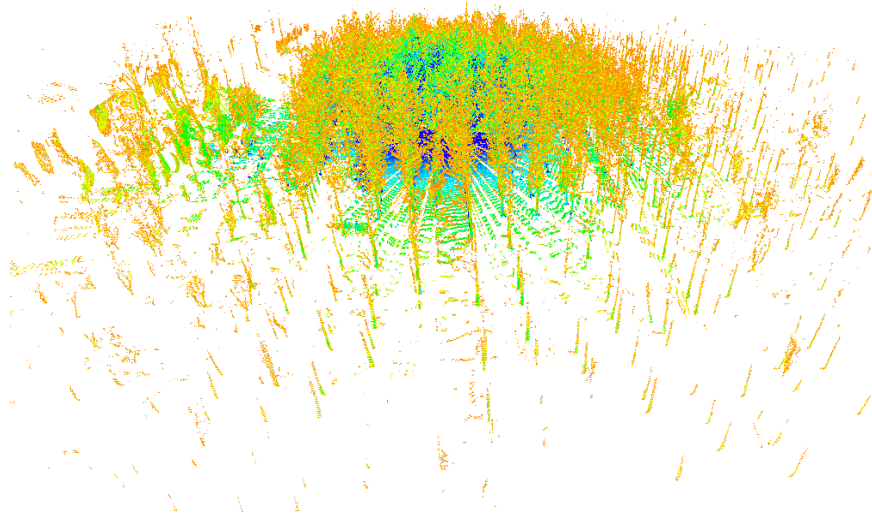

**Figure 6.** Point cloud data of a single scan acquired by using the BEE scanner.

In the study, four single scans were performed in the four plots, respectively, in the field experiments. In each single scan, a square plot with the size of $10\,m \times 10\,m$ was extracted from the whole single scan. The center point of the plot is the position of the BEE scanner. The point cloud data of plot 1 are shown in Figure 7, for example.

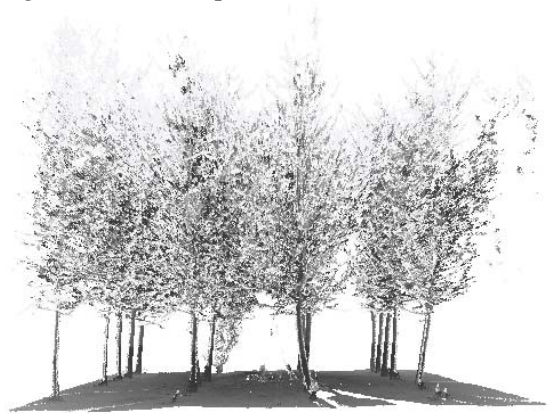

**Figure 7.** Point cloud data of plot 1.

Based on the dataset of each plot, the BEE scanner could obtain not only the point cloud data, but also the range and intensity data.

*4.2. Detecting Results*

According to the detecting method described in Section 3.2, the point cloud data in each plot was tested and some trees were detected. The detected results of four plots are shown in Figure 8. As shown in Figure 8, there is a group of points belonging to the local ground below each detected tree, and the ground points were used to fit the ground plane by using the method mentioned in Section 3.2.

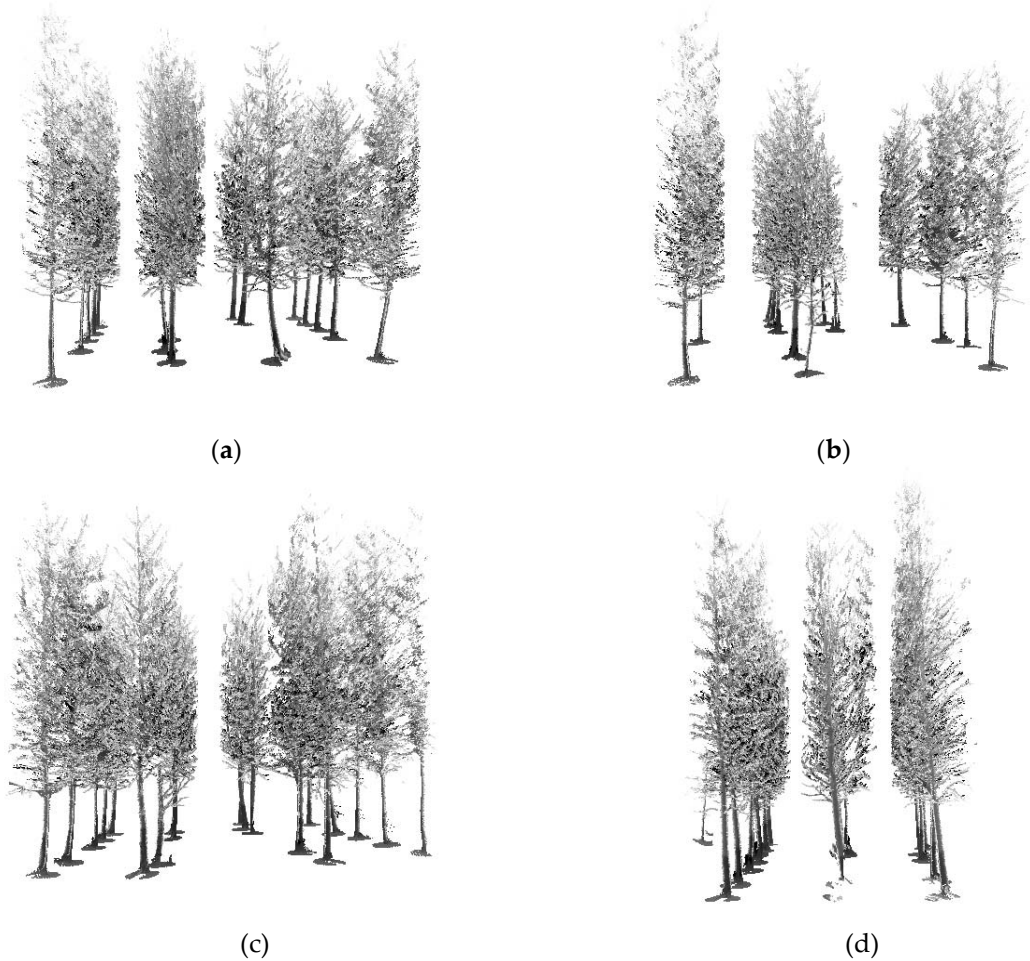

**Figure 8.** The detected trees of the four plots: (**a**) plot 1, (**b**) plot 2, (**c**) plot 3, and (**d**) plot 4.

The results of the trunk detection are reported in Table 2. The number of trees and the stem density were also listed in Table 2. As shown, the stem density was from 0.14 stems/m² to 0.22 stems/m². The correct detection means that the detected tree was in the right place. The false detection means that the detected tree did not exist, and an omission means that the tree existed but was not found.

The detection rate is the ratio of the correct detection and the number of trees. The highest detection rate was 100% in plot 1, and the lowest detection rate was 86.36% in plot 3. Table 2 shows that there are 69 trees in the four plots. Among these trees, 64 trees were detected and five trees were missed. Therefore, the overall detection rate was 92.75%. Among these missed trees, three trees did not appear in the point cloud data of the plot because of the shadowing effects. The other two trees were missed because of the detecting algorithm. The thresholds of the diameter and the fewer branches in the point cloud transect contributed to the zero value of false detection.

**Table 2.** The results of the trunk detection.

| Plot | Number of Trees | Correct detection | False detection | Omission | Stem density (stems/m²) | Detection rate |
|---|---|---|---|---|---|---|
| 1 | 16 | 16 | 0 | 0 | 0.16 | 100% |
| 2 | 14 | 13 | 0 | 1 | 0.14 | 92.86% |
| 3 | 22 | 19 | 0 | 3 | 0.22 | 86.36% |
| 4 | 17 | 16 | 0 | 1 | 0.17 | 93.75% |

*4.3. Estimating DBH*

Based on the results of the trunk detection, the DBH of the detected trees are estimated and the scatterplot of the tape-measured DBH and the estimated DBH are shown in Figure 9. The estimating error between the tape-measured DBH and the estimated DBH are mostly positive, accounting for 76.56%. That is to say, in general, the DBH was overestimated in the experiment.

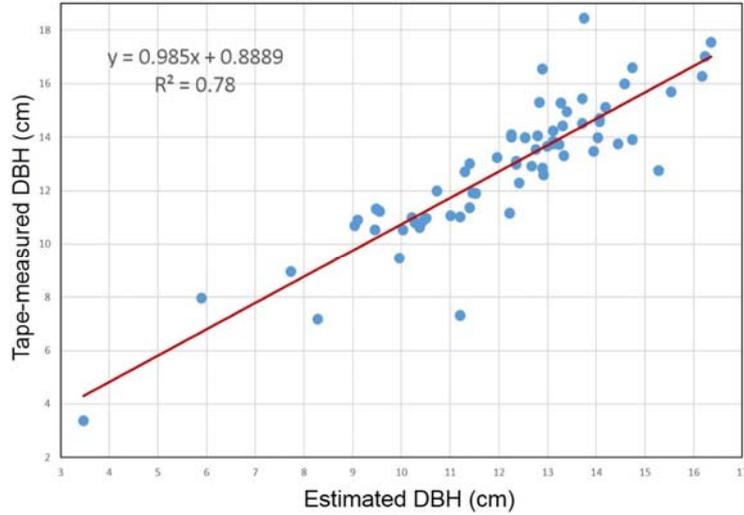

**Figure 9.** The scatterplot of the tape-measured DBH and the estimated DBH.

The bias and the RMSE values of the DBH estimating results were listed in Table 3. The bias and RMSE were calculated as described in Liang's paper [25].

**Table 3.** The results of the trees detection.

| Plot | Bias | Bias% | RMSE | RMSE% |
|---|---|---|---|---|
| DBH(cm) | 0.64 | 5.32 | 1.27 | 10.52 |

*4.4. Estimating Heights and Positions*

According to the results of the trunk detection and the local ground plane fitting, the point cloud data in the cylinder space around the tree was extracted. In the experiment, the diameter of the cylinder was 0.8 m, and the ground plane fitting was carried out based on the point cloud data with the lowest height of about 10 cm and with a diameter of 0.6 m. The local ground is relatively flat and the DBH of the tree is below 20 cm. The inclination angles of the fitted local ground planes are about 3 degrees, and the related height errors are about 1 cm. Therefore, the tilted ground has little effect on the DBH estimation.

A tree with the cylinder is shown in Figure 10, the red point $p_{zero}$ denotes the local zero-height point, and the green point $p_{top}$ denotes the uppermost point of the tree. Then the tree height could be calculated and the top point cloud data with a 0.5 m height, which was used to determine the point $p_{top}$, is also shown in Figure 10.

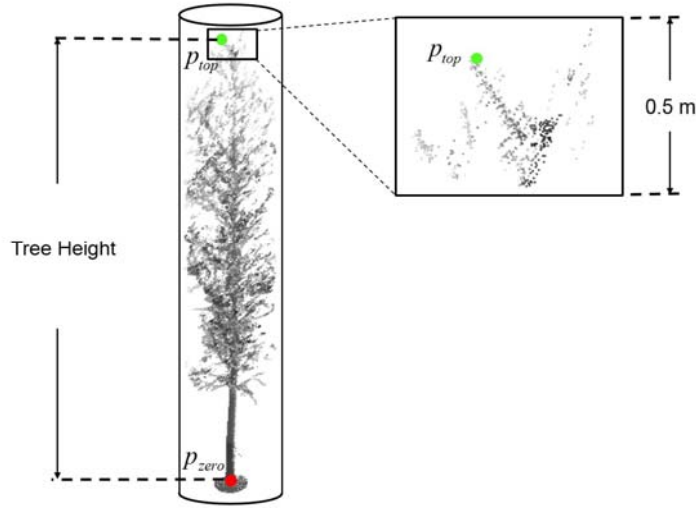

**Figure 10.** The demonstration of tree height estimation.

The $p_{top}$ is defined by using the clustering method described in Section 3.4. The top point cloud data shown in Figure 10 was used to demonstrate the procedure of clustering, which is shown in Figure 11. The ten nearest neighbor points of each point in the top point cloud data were found by using the KNN algorithm, and the green star in Figure 11a was an example point whose ten nearest neighbor points were plotted in red. The ten nearest neighbor points of each point were found in this way and were used to cluster the points into two clusters. One cluster are the tree points, which are plotted in blue in Figure 11b, and the other cluster are the outliers and the points that could not be definitely treated as tree points. These points are plotted in red. The uppermost point of the tree is defined as the highest point in the cluster of tree point $p_{top}$.

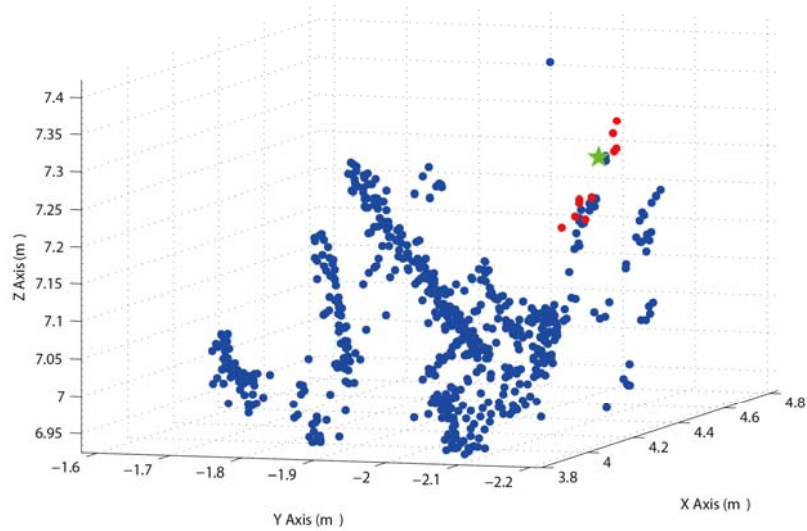

(a)

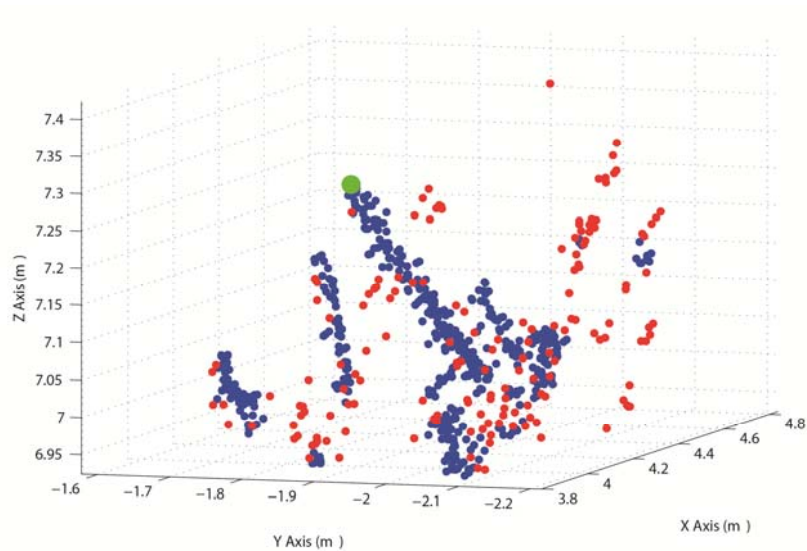

(b)

**Figure 11.** The process of the uppermost point definition. (**a**) An example point with its ten nearest neighbors. The green star is the example point and the red points are its ten nearest neighbors. (**b**) Two point clusters and the uppermost point. The cluster of tree points are plotted in blue. The outliers and other points are plotted in red, and the uppermost point of the tree is plotted in green with a larger size.

Tree height was estimated for each detected tree, and the scatterplot of the manual measured tree height and the estimated tree height is shown in Figure 12.

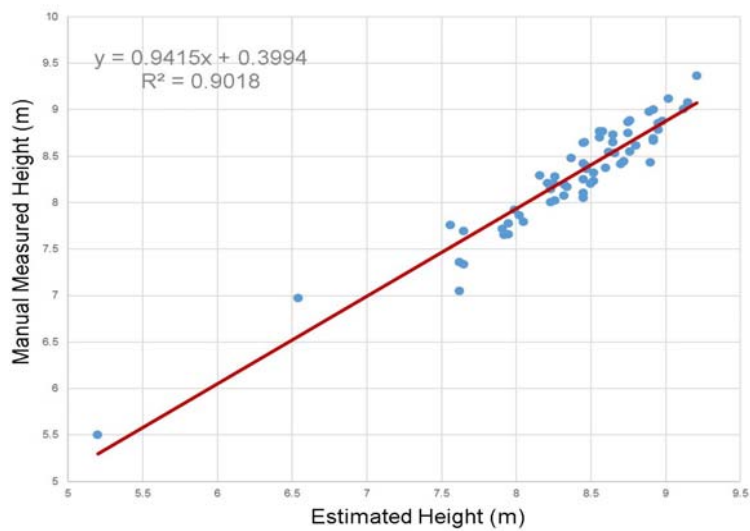

**Figure 9.** The scatterplot of the tape-measured DBH and the estimated DBH.

The bias and the RMSE values of the tree height estimation results are listed in Table 4. In the four experimental plots, the ginkgo trees are artificially planted at the same period. Most of the ginkgo trees are about 8 m high. In the experiment, the tree heights were obviously underestimated.

**Table 4.** The results of the trees detection.

| Plot | Bias | Bias% | RMSE | RMSE% |
|---|---|---|---|---|
| Height(m) | −0.08 | −0.92 | 0.24 | 2.77 |

The manual measured tree positions and the estimated tree positions of the four plots are shown in Figure 13. The blue diamonds in Figure 13 represent the position of the BEE scanner. The red circles represent the manual measured tree positions and the blue circle represent the estimated tree positions. The estimated tree positions are in good agreement with the manual measured positions.

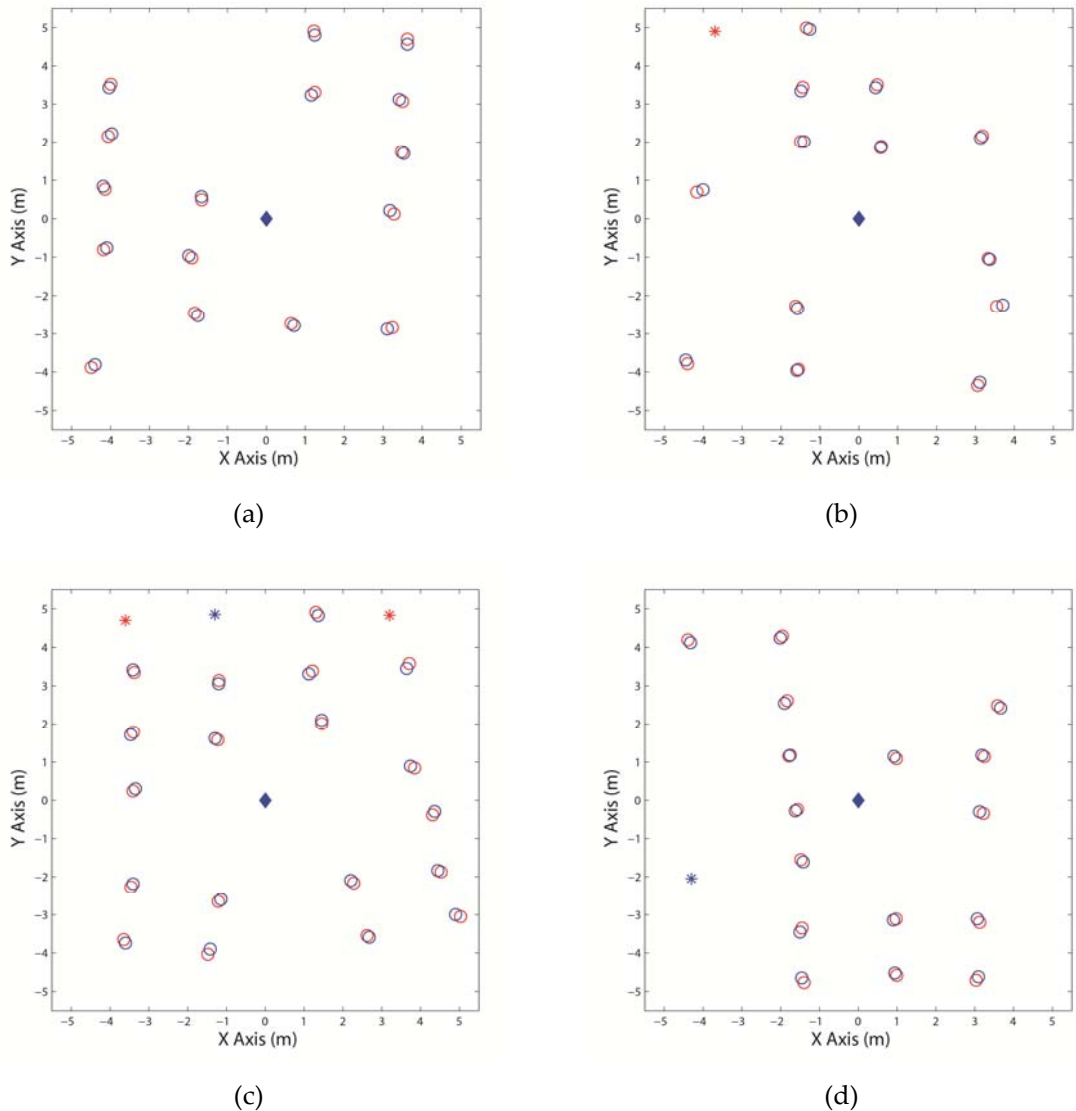

**Figure 13.** The map of tree positions for manual measured results (red) and estimated results (blue). (**a**) Plot 1; (**b**) Plot 2; (**c**) Plot 3; and (**d**) Plot 4.

## 5. Discussion

In this paper, we designed and implemented a low-cost 3D laser scanner (the BEE scanner) and also developed the supporting software, BEEScan, to capture the point cloud information of trees. The BEE scanner is portable and automated, and designed for forest inventory applications. Considering the performance of the laser scanner, the BEE scanner was designed for scanning small plots and estimating the tree parameters, such as DBH, tree height, and tree position. The BEE scanner could also be used to scan a larger area forest by multiple single scans. In the field measurement, the BEE scanner could give results quickly, which could be verified on site.

In this study, the correct detection rate is 92.75%, which is a high value. There are several reasons attributing to the higher correct detection rate. One reason is the smaller size of the plot. The BEE scanner is placed at the middle point in a plot which is $10m \times 10m$ in size. This ensures that the point cloud density is high enough to describe the trees in relative detail. The other reason is the smaller angular step-width in the horizontal direction, which means that points used to detect trees would be of a high density to improve the detection rate and the accuracy of DBH estimation. There is another reason that the lower stem density, which is 0.1725 stems/m$^2$, and the regular inter-tree spacing are good help to reduce the shadowing effects. The shadowing effect is one of the common problems in forest scanning, even in our small-scale experimental plots. However, the small size of the plot could reduce the passive impact to a certain extent.

The preset thresholds were used in the trunk detection to avoid the obvious false estimation that could happen in several circumstances. In the practical circle fitting, we encountered two cases which could obtain a result of DBH estimation, but the deviation from the real data is very large. In the first case, a small part of the trunk was scanned because of the shadowing effect. The point cloud transect was projected into the plane and the points were distributed in a small arc. In this case, the estimated DBH may be a very large value, which is obviously unreasonable. In the second case, there are some points of the lower branches and noise in the point cloud transect. If the distance between them matched our clustering method, they would be considered as the trunk points. The circle fitting based on these points could also obtain an unreasonable result. Therefore, the thresholds of the DBH value were used as a simple test which could reduce in false detection. However, it should be noted that a reasonable estimated DBH value may also be obtained in both of the above mentioned cases.

There are three missing trees because of the shadowing effects. The three trees are on the border of the plots and blocked by other trees. In addition to the three trees, two trees were missed because of the above mentioned preset thresholds in the method of trunk detection. One ginkgo tree was not detected because it is too close to a pine tree. The point cloud transect contains the ginkgo tree and the pine tree, and these points match our clustering method proposed in the paper, so they were thought to belong to the same tree. However, the circle fitting result based on these points is beyond the preset threshold, so the tree was missed. The other ginkgo tree was not detected because there are some branches and outliers in the point cloud transect, and the circle fitting result is beyond the preset threshold, too, so the tree was also missed.

To simplify the process and ensure the efficiency, the geometric circle fitting method of Pratt was used to detect trunks based on a single scan and a single slice, and the circle fitting method also provides the estimation of the DBH and tree positions. The experimental results of the DBH estimation shows that the RMSE is 1.27 cm and the bias is 0.64 cm in the study. The accuracy of the DBH estimation is reasonable and acceptable. Since the BEE scanner is composed of low-cost sensors and the accuracy of the sensor is not as good as expensive terrestrial laser scanners, and the BEE scanner was assembled of several parts, the accuracy of the whole system is not only affected by the sensor, but also by the other parts. Thus, there are many sources for estimating errors which will be analyzed in detail in our future work. However, the estimating results of DBH are still in line with the analysis by Pueschel [33] and Bu [28], in that the DBH estimation accuracy is not sensitive to the scanner parameter settings, but sensitive to the coverage of tree stems and point density. When the points used to estimate are fewer, the result of the DBH estimation is seriously affected by several individual outliers.

The estimating method of tree height used in the paper selected the uppermost point of a tree in the clustered tree points. The KNN method and an adaptive threshold were used in clustering. The proposed clustering method put the outliers and the points with low confidence into one cluster. The vertical angular step-width of the BEE scanner is relatively larger compared to the expensive scanners, and the point density is lower in the vertical direction. Thus, the tree tops were likely scanned incompletely because of the blocking and small size of sparse branches. Therefore, some points belonging to thin twigs and branches would not be clustered into tree points because of the sparse distribution and incomplete scanning. This will lead to the underestimation of tree heights. In addition to the DBH and tree heights, the tree positions were also estimated in the experiments. The result of the tree position estimation was nearly in line with the actual positions because of the small size of plots, the flat ground, and the straight trunks.

In addition to the plot-level estimating results mentioned above, the BEE scanner can also provide the range map and the intensity map, which could be used as auxiliary data for the processing and analysis in more applications. The BEE scanner has the advantages of low-cost and light weight, which will make it more practical and able to be used widely in the field of forest inventory. With the development of the laser sensors, the cost of the BEE scanner could be decreased further, the performance of the BEE scanner could be increased further, and more analysis methods could be designed and tested based on the dataset acquired by the scanner.

Compared with high cost of commercial 3D scanners, the BEE scanner has obvious disadvantages in precision and accuracy. According to the manual of the SICK LMS511 [34], the beam divergence is 4.7 mrad in high-resolution mode. That is to say the footprint of the SICK LMS511 will be about 4.7 cm at the distance of ten meters. The high-cost scanners have smaller beam divergences. For example, the beam divergence of the Riegl VZ-400 is 0.35 mrad [35], which is smaller than one-tenth of the beam divergence of the SICK LMS511. Therefore, the BEE scanner is considered to be more suitable for small-scale sample plots in single-scan mode.

## 6. Conclusions

An automated low-cost scanner, named the BEE scanner, was introduced in this paper. The estimations of DBH, tree height, and tree position were carried out based on the data acquired by using the BEE scanner in four small plots. The estimating results have verified the feasibility of the BEE scanner. The analysis of the results show that the BEE scanner has the potential for practical use in many fields of forest inventory, and the low-cost scanner could be an efficient method for acquiring the point cloud data of a forest. The point cloud data could be used in more application fields of forest inventory, such as forest mapping, derivation of tree skeletons, biomass estimation, and so on.

More tests and experiments should be done to study and analyze the attributes and performances of the BEE scanner, such as the merging of the multi-scan data of the BEE scanner, which could be used to reduce the shadowing effect and improve the accuracy of estimation. The BEE scanner could be tested in more complex and natural environments, and the portability, miniaturization, and automation of the BEE scanner could be one focus of our research in the future. Another focus is the data processing of the BEE scanner, which is very important and interesting for us.

In general, the BEE scanner has a lot of room for improvement in terms of hardware selection, scanning scheme, processing methods, and measuring accuracy. This kind of low-cost scanner will have great potential in forest inventory applications.

## Acknowledgments

The authors gratefully acknowledge the financial support from the Beijing Natural Science Foundation (No. 6164037) and the Fundamental Research Funds for the Central Universities (Nos. 2015ZCQ-LY-02 and YX2015-08).

## Author Contributions

Pei Wang: hardware design, algorithm development, design of the field experiment, and data analysis. Guochao Bu: hardware design, algorithm development, field measurement, and data analysis. Ronghao Li: field measurement and data analysis. Rui Zhao: data analysis.

## Conflicts of Interest

The authors declare no conflict of interest.